\documentclass[letterpaper]{article}
\usepackage{aaai}
\usepackage{graphicx}
\usepackage{times}
\usepackage{helvet}
\usepackage{courier}
\begin{document}

\title{Autonomous Deep Learning: Incremental Learning of Denoising Autoencoder for Evolving Data Streams}
\author{Mahardhika Pratama\textsuperscript{*,1}, Andri Ashfahani\textsuperscript{*,2}, Yew Soon Ong\textsuperscript{*,3}, Savitha Ramasamy\textsuperscript{+,4} and Edwin Lughofer\textsuperscript{\#,5}\\
\textsuperscript{*}School of Computer Science and Engineering, NTU, Singapore\\
\textsuperscript{+}Institute of Infocomm Research, A*Star, Singapore\\
\textsuperscript{\#}Johannes Kepler University Linz, Austria\\
\{\textsuperscript{1}mpratama@, \textsuperscript{2}andriash001@e., \textsuperscript{3}asysong@\}ntu.edu.sg, \textsuperscript{4}ramasamysa@i2r.a-star.edu.sg,\\ \textsuperscript{5}edwin.lughofer@jku.at\\
}
\copyrighttext{}
\maketitle
\begin{abstract}
\begin{quote}
The generative learning phase of Autoencoder (AE) and its successor Denosing Autoencoder (DAE) enhances flexibility of data stream method in exploiting unlabelled samples. Nonetheless, the feasibility of DAE for data stream analytic deserves in-depth study because it characterizes a fixed network capacity which cannot adapt to rapidly changing environments. An automated construction of a denoising autoeconder, namely deep evolving denoising autoencoder (DEVDAN), is proposed in this paper. DEVDAN features an open structure both in the generative phase and in the discriminative phase where input features can be automatically added and discarded on the fly. A network significance (NS) method is formulated in this paper and is derived from the bias-variance concept. This method is capable of estimating the statistical contribution of the network structure and its hidden units which precursors an ideal state to add or prune input features. Furthermore, DEVDAN is free of the problem- specific threshold and works fully in the single-pass learning fashion. The efficacy of DEVDAN is numerically validated using nine non-stationary data stream problems simulated under the prequential test-then-train protocol where DEVDAN is capable of delivering improvement of classification accuracy to recently published online 
learning works while having flexibility in the automatic extraction of robust input features and in adapting to rapidly changing environments. 
\end{quote}
\end{abstract}

\section{Introduction}
The underlying challenge in the design of DNNs is seen in the model selection phase where no commonly accepted methodology exists to configure the structure of DNNs \cite{DeepExpandable}. This issue often forces one to blindly choose the structure of DNNs. DNN model selection has recently attracted intensive research where the goal is to determine an appropriate structure for DNNs with the right complexity for given problems. It is evident that a shallow NN tends to converge much faster than a DNN and handles the small sample size problem better than DNNs. In other words, the size of DNNs strongly depends on the availability of samples. This encompasses the development of pruning  \cite{LearningTheNumber}, regularization \cite{parameterprediction}, parameter prediction \cite{parameterprediction}, etc. Most of which start with an over-complex network followed by a complexity reduction scenario to drop the inactive components of DNNs \cite{distilling}. These approaches, however, do not fully fit to handle streaming data problems because they rely on an iterative parameter learning scenario where the tuning phase is iterated across a number of epochs \cite{GamaDataStream}. Moreover, a fixed structure is considered to be the underlying bottleneck of this model because it does not embrace or is too slow to respond to new training patterns as a result of concept change especially if network parameters have converged to particular points \cite{GamaDataStream}. 

The ideas of online DNNs have started to attract research attention \cite{DEEPIOT}. In \cite{Zhou_incrementallearning}, online incremental feature learning is proposed using a denoising autoencoder (DAE) \cite{VincentDAE}. The incremental learning aspect is depicted by its aptitude to handle the addition of new features and the merging of similar features. The structural learning scenario is mainly driven by feature similarity and does not fully operate in the one-pass learning mode. \cite{OnlineDeepLearning} puts forward the hedge backpropagation method  to answer the research question as to how and when a DNN structure should be adapted. This work, however, assumes that an initial structure of DNN exists and is built upon a fixed-capacity network. To the best of our knowledge, the two approaches are not examined with the prequential test-then-train procedure considering the practical scenario where data streams arrive without labels, thus being impossible to first undertake the training process \cite{GamaDataStream}.

In the realm of DNNs, the pre-training phase plays a vital role because it addresses the random initialization problem leading to slow convergence \cite{Bengio_2013}. From Hinton’s variational bound theory \cite{HinSal06}, the  power of depth can be achieved provided the hidden layer has sufficient complexity and appropriate initial parameters. An unsupervised learning step is carried out in the pre-training phase, also known as the generative phase \cite{HinSal06}. The generative phase implements the feature learning approach which produces a higher-level representation of the input features and induces appropriate intermediate representation \cite{Hinton_AE}. From the viewpoint of data stream, the generative phase offers refinement of predictive model with the absence of true class label. This case is evident due to the fact that data stream often arrives without labels. Of the several approaches for the generative phase, the autoencoder (AE) is considered the most prominent method \cite{Hinton_AE}. DAE is a variant of AE which adopts the partial destruction of the original input features \cite{VincentDAE}. This approach prevents the learning identity function problem and opens the manifold of the original input dimension because the destroyed input variables are likely to sit further than the clean input manifold. Nevertheless, the structure of DAE is user-defined and not well-suited for data stream applications due to their iterative nature. 

A deep evolving denoising autoencoder (DEVDAN) for evolving data streams is proposed in this paper. DEVDAN presents an incremental learning approach for DAE which features a fully open and single-pass working principle in both generative and discriminative phase. It is capable of starting its generative learning process from scratch without an initial structure. Its hidden nodes can be automatically generated, pruned and learned on demand and on the fly. Note that this paper considers the most challenging case where one has to grow the network from scratch but the concept is directly applicable in the presence of initial structure. The discriminative model relies on a soft-max layer which produces the end-output of DNN and shares the same trait of the generative phase: online and evolving. DEVDAN distinguishes itself from \cite{Zhou_incrementallearning} because it works by means of estimation of network significance leading to approximation of bias and variance and is free of user-defined thresholds. A new hidden unit is introduced if the current structure is no longer expressive enough to represent the current data distribution - underfitting whereas an inconsequential unit is pruned in the case of high variance - overfitting. In addition, the evolving trait of DEVDAN is not only limited to the generative phase but also the discriminative phase.

The unique feature of the NS measure is its aptitude  to estimate the statistical contribution of a neural network and a hidden node during their lifespan in an online fashion.  This approach is defined as a limit integral  representation of a generalization error which approximates both the historical and future significance of the overall network and its hidden unit. It is worth mentioning that a different approach from conventional self-organizing radial basis function networks \cite{Platt,MRAN} has to be  developed  because  DAE  cannot  be  approached by an input space clustering method. The NS method offers a general framework of a statistical contribution measure and is extendable for different DNNs. Moreover, the NS method is also free of user-defined parameters which are often problem-dependent and hard to assign. It is supported by an adaptive conflict threshold dynamically adjusted with respect to the true performance of DEVDAN and current data distribution. 

The performance of DEVDAN has been numerically investigated using nine prominent data stream problems: SEA \cite{SEA}, Hyperplane \cite{MOA}, HEPMASS, SUSY \cite{Baldi2014SearchingFE}, KDDCup \cite{KDDCup}, Weather, electricity pricing \cite{DitzlerImbalanced}, RLCPS \cite{RLCPS}, RFID localization problem. DEVDAN is capable of improving accuracy of conventional DAE and outperforming proposed data stream methods \cite{pENsemble,pensembleplus}. It offers a flexible approach to the automatic construction of robust features from data streams and operates in the   one-pass learning fashion. Our numerical results are produced under \textbf{the prequential test-then-train protocol} - standard evaluation procedure of data stream method \cite{GamaDataStream}. The remainder of this paper is structured as follows: this paper starts with the problem formulation followed by the automatic construction of network structure and the discriminative training phase. The proof of concepts discusses numerical study in nine data stream problems and comparison of DEVDAN against state-of-the art algorithms. Some concluding remarks are drawn in the last section of this paper.  

\section{Problem Formulation}
Evolving data streams refer to continuous arrival of data points $B_k=[B_1,B_2,...,B_K]$ in a number of time stamps $K$ where $B_k$ may consist of a single data point $B_k=X_1\in\Re^{n}$ or be formed as a data batch of a particular size $B_k=[X_1,X_2,X_t,...,X_T]\in\Re^{T\times n}$. $n$ here denotes the input space dimension and $T$ stands for the size of data chunk. The size of data batch often varies and the number of time stamps is in practise unknown. In realm of real data stream environments, data points come into picture with the absence of true class labels $C\in\Re^{T}$. Labelling process is carried out and is subject to the access of ground truth or expert knowledge \cite{GamaDataStream}. In other words, a delay is expected in consolidating the true class labels. This issue warrants a generative learning step which can be applied to refine a predictive model in a unsupervised fashion while pending for operator to annotate the true class label of data samples - \textbf{the underlying motivation} of DEVDAN's algorithmic development. This problem also hampers the suitability of the conventional cross validation method or the direct train-test partition method as an evaluation protocol of data stream learner. Hence, the so-called \textbf{prequential test-then-train} procedure is carried out here. That is, data streams are first used to test the generalization power of a learner before being exploited to perform model's update. The performance of a data stream method is evaluated by aggregation of its performance across all time stamps.   

DEVDAN is constructed under the denoising autoencoder (DAE) \cite{VincentDAE} - a variant of autoencoder (AE) \cite{Hinton_AE} which aims to retrieve the original input information $X_t$ from the noise perturbation. The masking noise scenario is chosen here to induce partially destroyed input feature vector $\widetilde X_t$ by forcing its $n'$ elements to zeros. In other words, only a subset of original input features $n-n'$ goes through DAE. $n'$ corrupted input variables are randomly destructed in every training observation satisfying the joint distribution $q(\widetilde X,X)$ \cite{VincentDAE}. This mechanism brings DAE a step forward of classical AE since it never functions as an identity function rather extracts key features of predictive problem. The reconstruction process is carried out via encoding-decoding scheme formed with the sigmoid activation function $\frac{1}{1+exp(-s)}$ as follows:
\begin{equation}
y=f_{(W,b)}=s(\widetilde X_t W+b)\label{encoder}
\end{equation}
\begin{equation}
z=f_{(W',c)}=s(y W'+c)\label{decoder}
\end{equation}
where $W\in\Re^{n \times R}$ is a weight matrix, $b\in\Re^{R},c\in\Re^{n}$ are respectively the bias of hidden units and the decoding function. $R$ is the number of hidden units. The weight matrix of the decoder is constrained such that $W'$ is a reverse mapping $W^{T}$. That is, DAE has a tied weight \cite{VincentDAE}.

The typical characteristic of data stream is the presence of concept drift formulated as a change of the joint-class posterior probability $P(Y_t,X_t)\neq P(Y_{t-1},X_{t-1})$ \cite{Gamaconceptdrift}. This situation leads to a current model created by previously induced concept $B_{k-1}$ being obsolete. DEVDAN features an open structure where it is capable of initiating its structure from scratch without the presence of a pre-configured structure. Its structure automatically evolves in respect of the network significance approach forming an approximation of the network bias and variance. In other words, DEVDAN initially extracts a single input feature $R=1$ where the number of extracted input features incrementally augments $R=R+1$ if it signifies a underfitting situation, high bias, or decreases $R=R-1$ if it suffers from an overfitting situation, high variance. In realm of concept drift, this is supposed to handle the so-called virtual drift - distributional change of the input space. The virtual drift is interpreted by the change of prior probability $P(X)$ or the class conditional probability $P(X|Y)$ \cite{Gamaconceptdrift}. The parameter tuning scenario is driven by the stochastic gradient descent (SGD) method in \textbf{a single pass} mode with the cross-entropy cost function \cite{Bengio_Greedy}. 

Once the true class labels of a data batch $B_k$ has been observed $C_k$, the 0-1 encoding scheme is undertaken to construct a labelled data batch $(X_k,C_k)\in\Re^{T\times (n+m)}$  where $m$ stands for the number of target classes. The discriminative phase of DEVDAN is carried out once completing the generative phase of DEVDAN using a softmax layer trained with the SGD method with only \textbf{a single epoch}. Furthermore, the discriminative training process is also equipped by the hidden unit growing and pruning strategies derived in a similar manner as that of the generative training process. An overview of DEVDAN's learning mechanism is depicted in Fig. 1. One must bear in mind that DEVDAN's learning scheme can be also applied with an initial model.  
\begin{figure*}[t!]
	\begin{centering}
	\includegraphics[scale=0.45]{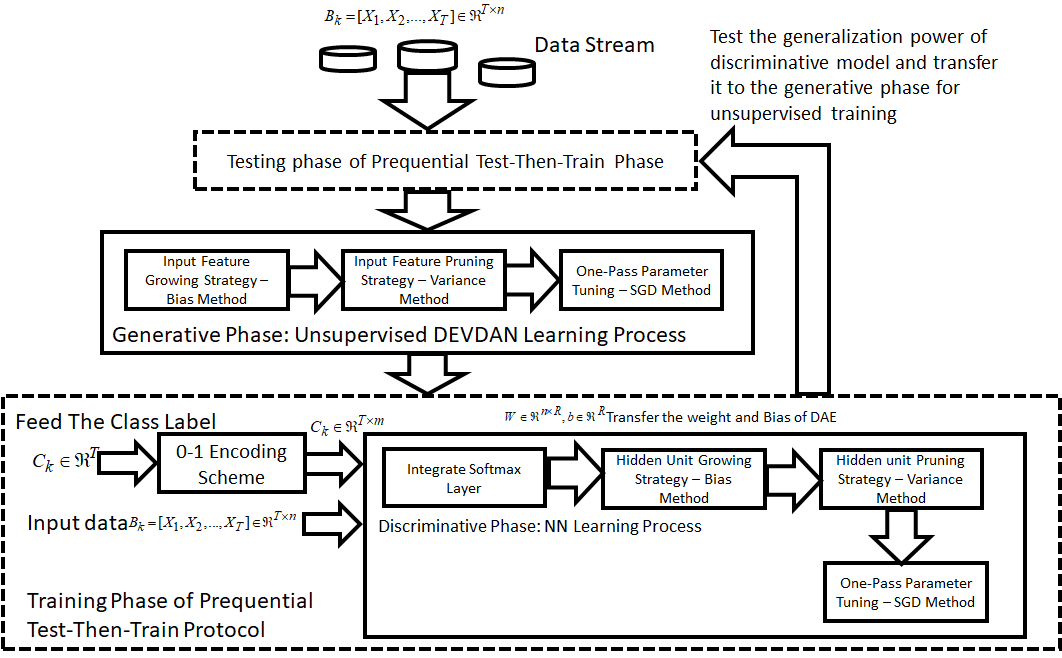}
	\par\end{centering}
	\caption{Learning Mechanism of DEVDAN}
	\label{fig:PHO}
\end{figure*}

\section{Automatic Construction of Network Structure}
This section formalizes the network significance (NS) method applied to grow and to prune hidden units of DAE.  

\subsection{Growing Hidden Units of DAE}\label{AA}
 The power of DAE can be examined from its reconstruction error which can be formed in terms of mean square error (MSE) as follows:
\begin{equation}
MSE=\sum_{t=1}^{T}\frac{1}{T}(X_t-z_t)^{2}    
\end{equation}
where $X_t,z_t$ respectively stand for clean input variables and reconstructed input features of DAE. This formula suffers from two bottlenecks for the single-pass learning scenario: 1) it calls for memory of all data points to understand a complete picture of DAE's reconstruction capability; 2) Notwithstanding that the MSE can be calculated recursively without revisiting preceding samples, this procedure does not examine the reconstruction power of DAE for unseen data samples. In other words, it does not take into account the generalization power of DAE. 

To correct this drawback, let $z$ denotes the estimation of clean input variables $x$ and $E[z]$ stands for the expectation of DAE's output, the NS method is defined as follows:
\begin{equation}
    NS=\int_{-\infty}^{\infty}(x-z)^{2}p(x)dx
    \end{equation}
Note that $E[x]=\int_{-\infty}^{\infty}xp(x)dx$ where $p(x)$ is the probability density estimation. The NS method can be defined in terms of the expectation of the squared reconstruction error:
    \begin{equation}
    NS=E[(x-z)^{2}]=E[(z-E[z]+E[z]-x)^{2}]
\end{equation}
Several mathematical derivation steps lead to the bias and variance formula as follows:
\begin{equation}
NS=E[(z-E[z])^{2}]+(E[z]-x)^{2}=Var(z)+Bias(z)^{2}\label{eq:NS}    
\end{equation}
where the variance of a random variable $z$ can be expressed as $Var(z)=E[(z-E[z])^{2})]=\int_{-\infty}^{\infty}(z-E[z])^{2}p(x)dx=E[z^{2}]-E[z]^{2}$. The key for solving (\ref{eq:NS}) is to find the expectation of the recovered input attributes delineating the statistical contribution of DAE. It is worth mentioning that the statistical contribution captures both the network contribution in respect to past training samples and unseen samples. It is thus written as follows:

\begin{equation}
E[z]=\int_{-\infty}^{\infty}s(yW'+c)p(y)dy\label{eq:Ez}
\end{equation}
It is evident that $y$ is induced by the feature extractor $s(\widetilde x +b)$ and is influenced by partially destroyed input features $\widetilde x$ due to the masking noise. Hence, (\ref{eq:Ez}) is modified as follows:
\begin{equation}
    E[z]=s(E[y]W'+c)
\end{equation}
\begin{equation}
    E[y]=\int_{-\infty}^{\infty}s(\widetilde x W+b)p(\widetilde x)d\widetilde x \label{eq:E_y}
\end{equation}
Suppose that the normal distribution holds, the probability density function (PDF) $p(\widetilde x)$ is expressed as $\frac{1}{\sqrt{2\pi}}exp(-\frac{(\widetilde x - \mu)^{2}}{\sigma^{2}})$. It is also known that the sigmoid function can be approached by the probit function $\Phi(\xi x)$ \cite{Murphy_Machine_Learning} where $\Phi(x)=\int_{-\infty}^x \mathcal{N}(\theta|0,1)d\theta$ and $\xi^{2}=\pi/8$. Following the result of \cite{Murphy_Machine_Learning}, (\ref{eq:E_y}) is derived:
\begin{equation}
    E[y]=s(\frac{\mu}{\sqrt{1+\pi \sigma^{2}/8}}W+b)
\end{equation}
where $\mu,\sigma$ are respectively the mean and standard deviation of the Gaussian function which can be calculated recursively from streaming data. The final expression of $E[z]$ is formulated as follows:
\begin{equation}
E[z]=s(s(\frac{\mu}{\sqrt{1+\pi \sigma^{2}/8}}W+b)W'+c) \label{eq:E_z2}
\end{equation}
where (\ref{eq:E_z2}) is a function of two sigmoid functions. This result enables us to establish the $Bias^{2}=(E[z]-x)^{2}$ in (\ref{eq:NS}). Let's recall $var(z)=E[z^{2}]-E[z]^{2}$. The second term $E[z]^{2}$ is derived from (\ref{eq:E_z2}) while the first term $E[z^{2}]$ is written:
\begin{equation}
    E[z^{2}]=s(E[y^{2}]W'+c)
   \end{equation}
   Due to the fact that $y^{2}=y*y$ , it is obvious that $y^{2}$ is IID variable which allows us to go further as follows:
  \begin{equation}
    E[z^{2}]=s(E[y]E[y]W'+c)
   \end{equation}  
     \begin{equation}
    E[z^{2}]=s(s(\frac{\mu}{\sqrt{1+\pi \sigma^{2}/8}}W+b)^{2}W'+c) \label{eq:Ez^}
   \end{equation}  
Consolidating all the results of (\ref{eq:E_z2}) and (\ref{eq:Ez^}), the final expression of the NS method is established. The NS method is derived from the expectation of MSE leading to the popular bias and variance formula. This method allows one to examine the quality of the predictive model by directly inspecting the possible underfitting or overfitting situation of a predictive model and capturing the reliability of a predictive model across the overall data space given a particular data distribution. A high NS value indicates either a high variance problem (overfitting) or a high bias problem (underfitting) which cannot be simply portrayed by a system error index. The addition of a new hidden node is supposed to reduce the high bias problem. It is, however, not to be done in the case of overfitting because it exacerbates the overfitting situation.

The hidden unit growing condition is derived from a similar idea to statistical process control which applies the statistical method to monitor the predictive quality of DEVDAN and does not rely on the user-defined parameter \cite{Gama2006,Gamaconceptdrift}. Nevertheless, the hidden node growing condition is not modelled as the binomial distribution here because DEVDAN is more concerned about how to reconstruct corrupted input variables rather than performing binary classification. Because the underlying goal of the hidden node growing process is to relieve the high bias problem, a new hidden node is added if the following condition is satisfied:
\begin{equation}
    \mu_{Bias}^{t} + \sigma_{Bias}^{t} \geq \mu_{Bias}^{min} + \pi\sigma_{Bias}^{min}\label{eq:HUgrowing}
\end{equation}
where  $\mu_{Bias}^{t},\sigma_{Bias}^{t}$ are respectively the mean and standard deviation of Bias at the $t-th$ time instant while $\mu_{Bias}^{min},\sigma_{Bias}^{min}$ are the minimum Bias up to the $t-th$ observation. These variables are computed with the absence of previous data samples by simply updating their values whenever a new sample becomes available. Moreover, $\mu_{Bias}^{min},\sigma_{Bias}^{min}$ have to be reset once (\ref{eq:HUgrowing}) is satisfied. Note that the bias can be calculated by decomposing the NS formula in (\ref{eq:NS}).  This setting is also formalized from the fact that the Bias values should decrease while the number of training observations increases as long as there is no change in the data distribution. On the other hand, a rise in the Bias values signals the presence of concept drift which cannot be addressed by simply learning the DAE's parameters. A similar approach is adopted in the drift detection method (DDM) \cite{Gama2006} but no warning phase is arranged in the NS method to avoid the use of windowing approaches. (\ref{eq:HUgrowing}) is derived from the so-called sigma rule where $\pi$ governs the confidence degree of sigma rule. $\pi$ is selected as $1.3exp(-{bias^2})+0.7$ which leads $\pi$ to revolve around $[1,2]$ meaning that it attains the confidence level of 68.2\% to 95.2\%. This strategy aims to improve flexibility of hidden unit growing process which adapts to the learning context and addresses the problem-specific nature of the constant $\pi$. A high bias signifies a underfitting situation which can be resolved by adding complexity of network structure while addition of hidden unit should be avoided in the case of low bias to prevent the variance increase.  

Once a new hidden node is appended, its parameters, $b$ is randomly sampled from the scope of $[-1,1]$ for simplicity while $W$ is allocated as $-e$. This formulation comes from the fact that a new hidden unit should drive the error toward zero. In other words, $e=X_t-s(y_tW'+c)+s_{R+1}(yW_{R+1}^{'}+c)=0$ where $R$ is the number of hidden units or extracted features. New hidden node parameters play crucial role to assure improvement of reconstruction capability and to drive to a zero reconstruction error. It is accepted that the scope $[-1,1]$ does not always ensure model's convergence. This issue can be tackled with adaptive scope selection of random parameters \cite{SCN}. 

\subsection{Hidden Unit Pruning Strategy}
The overfitting problem occurs mainly due to a high network variance resulting from an over-complex network structure. The hidden unit pruning strategy helps to find a lower dimensional representation of feature space by discarding its superfluous  components. Because a high variance designates the overfitting condition, the hidden unit pruning strategy starts from the evaluation of model's variance. The same principle as the growing scenario is implemented where the statistical process control method is adopted to detect the high variance problem as follows:
\begin{equation}
        \mu_{Var}^{t} + \sigma_{Var}^{t} \geq \mu_{Var}^{min} + 2\chi\sigma_{Var}^{min}\label{eq:HUpruning}
\end{equation}
where $\mu_{Var}^{t},\sigma_{Var}^{t}$ respectively stand for the mean and standard deviation of $Var$ at the $t-th$ time instant while $\mu_{Var}^{min},\sigma_{Var}^{min}$ denote the minimum Bias up to the $t-th$ observation. $\chi$, selected as $1.3exp(-{Var})+0.7$, is a dynamic constant controlling the confidence level of the sigma rule. The term 2 is arranged in (\ref{eq:HUpruning}) to overcome a direct-pruning-after-adding problem which may take place right after the feature growing process due to the temporary increase of network variance. The network variance naturally alleviates as more observations are encountered. Note that $Var$ can be calculated with ease by following the mathematical derivation of the NS method. Moreover, $\mu_{Var}^{min},\sigma_{Var}^{min}$ are reset when (\ref{eq:HUpruning}) is satisfied.  

After (\ref{eq:HUpruning}) is identified, the contribution of each hidden unit is examined. Inconsequential hidden unit is discarded to reduce the overfitting situation. The significance of hidden unit is tested via the concept of network significance, adapted to evaluate the hidden unit statistical contribution. This method can be derived by checking the hidden node activity in the whole corrupted feature space $\widetilde x$. The significance of the $i-th$ hidden node is defined as its average activation degree for all possible data samples as follows:
\begin{equation}
    HS_i=\lim_{T\to\infty}\sum_{t=1}^{T}\frac{s_i(\widetilde x W_i+b_i)}{T} \label{eq:NS_i2}
\end{equation}
where $W_i,b_i$ stand for the connective weight and bias of the $i-th$ encoding function. Suppose that data samples are sampled from a certain PDF, (\ref{eq:NS_i2}) can be derived as follows:
\begin{equation}
    HS_i=\int_{-\infty}^{\infty} s_i(\widetilde xW_i+b_i) p(\widetilde x) d \widetilde x \label{eq:NS_i3}
\end{equation}
Because the decoder is no longer used and is only used to complete a feature learning scenario, the importance of the hidden units is examined from the encoding function only. As with the growing strategy, (\ref{eq:NS_i3}) can be solved from the fact that the sigmoid function can be approached by the Probit function. The importance of the $i-th$ hidden unit is formalized as follows:
\begin{equation}
    HS_i=s(\frac{\mu}{\sqrt{1+\pi \sigma^{2}/8}}W_i+b_i)\label{eq:NS_expect}
\end{equation}
where $\mu,\sigma$ respectively denote the mean and standard deviation of the partially destroyed input features $\widetilde x$. Because the significance of the hidden node is obtained from the limit integral of the sigmoid function given the normal distribution, (\ref{eq:NS_expect}) can be also interpreted as the expectation of $i-th$ sigmoid encoding function. It is also seen that (\ref{eq:NS_expect}) delineates the statistical contribution of the hidden unit in respect to the recovered input attribute. A small HS value implies that $i-th$ hidden unit plays a small role in recovering the clean input attributes $x$ and thus can be ruled out without significant loss of accuracy.

Since the contribution of $i-th$ hidden unit is formed in terms of the expectation of an activation function, the least contributing hidden unit having the minimum $HS$ is deemed inactive. If the overfitting situation occurs or (\ref{eq:HUpruning}) is satisfied, the pruning process encompasses the hidden unit with the lowest $HS$ as follows:

\begin{equation}
 Pruning \longrightarrow \min_{i=1,...,R} HS_i \label{eq:pruning}
\end{equation}
The condition (\ref{eq:pruning}) aims to mitigate the overfitting situation by getting rid of the least contributing hidden unit. This condition also signals that the original feature representation can be still reconstructed with the rest of $R-1$ hidden units. Moreover, this strategy is supposed to enhance the generalization power of DEVDAN by reducing its variance. 

\subsection{Generative Training Phase}
 The parameter optimization phase is carried out using the stochastic gradient descent (SGD) approach with only single epoch. Since data points are normalized into the range of $[0,1]$ \cite{Bengio_Greedy}, the SGD procedure is derived using the cross-entropy loss function as follows:
\begin{equation}
W,b,c=\arg\min_{W,b,c}\sum_{t=1}^{T}\frac{1}{T}L(X_t,z_t)
\end{equation}
\begin{equation}
    L(X_t,z_t)=-\sum_{t=1}^{T}[X_t log(z_t)+(1-X_t) log(1-z_t)]
\end{equation}
where $X_t\in\Re^{n}$ is the noise-free input vector and $z_t\in\Re^{n}$ is the reconstructed input vector. $T$ is the number of samples observed thus far. Note that the cross-entropy function can be seen as the negative log-likelihood function. The minimization of the negative log-likelihood function is equivalent to the maximization of the likelihood function from the maximum likelihood optimization principle. Because the SGD method is utilized in the parameter learning scenario to update $W,b,c$, the tuning phase is carried out on a per-sample basis or a single-pass scenario. $T$ is thus set as 1. The first order derivative in the SGD method is calculated with respect to the tied weight constraint $W'=W^{T}$. Note that the parameter adjustment step is carried out under a dynamic network which commences with only a single input feature $R=1$ and grows its network structure on demand.

The notion of DEVDAN allows the model's structure to be self-organized in the generative phase while pending for operator to feed the true class labels $C_k$. Furthermore, the concept of DAE discovers salient structure of input space by opening manifold of learning problem and expedites parameter's convergence in the discriminative training phase. all of which can be committed while pending for operator to feed the true class labels. Although DEVDAN is realized in the single hidden layer architecture, it is modifiable to the deep structure with ease by applying the greedy layer-wise learning process \cite{Bengio_Greedy}. 
\begin{table*}[!t]
\caption{Numerical results of benchmarked algorithm}
\begin{centering}
\begin{tabular}{|l|l|r|r|r|r|r|}
\hline 
Data sets & Performance & DEVDAN & pEnsemble & pEnsemble+ & AE & DAE \tabularnewline
\hline 
\hline 
SUSY & CR & $\textbf{77.53}\pm\textbf{3.22}$ & $74.44\pm{2.4}$ & $76.99\pm{4.6}$ & $76.37\pm{3.91}$ & $76.24\pm{4}$\tabularnewline
\cline{2-7} 
 & ET & $4$K & $13$K & $35$K & $1$K & $1$\tabularnewline
\cline{2-7} 
 & HN & $20.6\pm{2.8}$ & $2.09\pm{0.99}$ & $8.94\pm{2.91}$ & $10$ & $10$ \tabularnewline
\cline{2-7} 
 & NoP & $435.5\pm{58.9}$ & $36.43\pm{21.21}$ & $230\pm{80}$ & $212$ & $212$ \tabularnewline
\hline 
HEPMASS & CR & $\textbf{83.91}\pm\textbf{2.45}$ & ${82.6}\pm{1.9}$ &  $82.3\pm{2.2}$ & $79.92\pm{2.73}$ & $79.8\pm{2.69}$ \tabularnewline
\cline{2-7} 
$19\%$ & ET & $1.2$K & $12$K & $7.6$K & $464.45$ & $519$ \tabularnewline
\cline{2-7} 
 & HN & $19.99\pm{0.4}$ & $2.01\pm{0.69}$ & $2.01\pm{0.69}$ & $10$ & $10$ \tabularnewline
\cline{2-7} 
 & NoP & $622.5\pm{15}$ & $24.14\pm{8.23}$ & $24.14\pm{8.23}$ & $312$ & $312$ \tabularnewline
\hline
RLCPS & CR & $\textbf{99.99}\pm\textbf{0.03}$ & $99.7\pm{0.3}$ & $99.8\pm{0.3}$ & $99.99\pm{0.03}$ & $99.99\pm{0.04}$ \tabularnewline
\cline{2-7} 
 & ET & $7$K & $60$K & $12.6$K & $1$K & $1$K \tabularnewline
\cline{2-7} 
 & HN & $60.23\pm{1.82}$ & $49.7\pm{15.14}$ & $6.96\pm{1.06}$ & $10$ & $10$ \tabularnewline
\cline{2-7} 
 & NoP & $724.68\pm{26.82}$ & $24$ & $83.52\pm{12.72}$ & $122$ & $122$ \tabularnewline
\hline
RFID & CR & $98.9\pm{3.33}$ & $60.4\pm{6.7}$ & $60.9\pm{7.6}$ & $99.02\pm{3.34}$ & $\textbf{99.19}\pm\textbf{2.03}$ \tabularnewline
\cline{2-7} 
 localization & ET & $176.58$ & $499$ & $700$ & $49.94$ & $60.81$ \tabularnewline
\cline{2-7} 
 & HN & $51.74\pm{9.47}$ & $1.57\pm{0.65}$ & $1.31\pm{0.46}$ & $10$ & $10$ \tabularnewline
\cline{2-7} 
 & NoP & $417.54\pm{76.94}$ & $42.7\pm{22.48}$ & $43.73\pm{13.52}$ & $84$ & $84$ \tabularnewline
 \hline
 Electricity & CR & $69.4\pm{8.74}$ & $\textbf{72.6}\pm\textbf{11.4}$ & $72.6\pm{12.1}$ & $67.72\pm{10.48}$ & $68.61\pm{8.55}$ \tabularnewline
\cline{2-7} 
 pricing & ET & $17.32$ & $71.2$ & $78.2$ & $8.18$ & $9.67$ \tabularnewline
\cline{2-7} 
 & HN & $10.58\pm{1.26}$ & $1$ & $1.01\pm{0.12}$ & $10$ & $10$ \tabularnewline
\cline{2-7} 
 & NoP & $117.74\pm{17.7}$ & $12$ & $12$ & $112$ & $112$ \tabularnewline
 \hline
 Weather & CR & $74.04\pm{5.68}$ & ${78.4}\pm{4.3}$ & $\textbf{78.8}\pm\textbf{4}$ & $73.76\pm{5.76}$ & $71.18\pm{7.06}$ \tabularnewline
\cline{2-7} 
 & ET & $6.93$ & $ 33.49 $ & $29.42$ & $3.2$ & $3.7$ \tabularnewline
\cline{2-7} 
 & HN & $14\pm{0.57}$ & $1$ & $1$ & $10$ & $10$ \tabularnewline
\cline{2-7} 
 & NoP & $153.54\pm{24.56}$ & $24$ & $24.33\pm{2}$ & $112$ & $112$ \tabularnewline
 \hline
 KDDCup & CR & $\textbf{99.84}\pm\textbf{0.21}$ & $99.3\pm{0.4}$ & $96.7\pm{6}$ & $99.83\pm{0.21}$ & $99.81\pm{0.21}$ \tabularnewline
\cline{2-7} 
 10\% & ET & $338.6$ & $5362.9$ & $860$ & $116.69$ & $133.37$ \tabularnewline
\cline{2-7} 
 & HN & $36.01\pm{6.09}$ & $1$ & $1$ & $10$ & $10$ \tabularnewline
\cline{2-7} 
 & NoP & $1587\pm{270.72}$ & $12$ & $12$ & $442$ & $442$ \tabularnewline
 \hline
 SEA & CR & $\textbf{92.29}\pm\textbf{6.48}$ & $92\pm{5.7}$ & $92\pm{6}$ & $91.74\pm{6.61}$ & $92.12\pm{6.34}$ \tabularnewline
\cline{2-7} 
 & ET & $38.88$ & $178.2$ & $200$ & $18.59$ & $21.4$ \tabularnewline
\cline{2-7} 
 & HN & $18.4\pm{10.4}$ & $2.51\pm{0.81}$ & $2.51\pm{0.81}$ & $10$ & $10$ \tabularnewline
\cline{2-7} 
 & NoP & $112.35\pm{62.87}$ & $60.3\pm{19.43}$ & $60.3\pm{19.43}$ & $62$ & $62$ \tabularnewline
 \hline
 Hyperplane & CR & $\textbf{92.12}\pm\textbf{3.47}$ & $91.8\pm{1.9}$ & $87.6\pm{6.2}$ & $90.92\pm{3.18}$ & $91.43\pm{3.29}$ \tabularnewline
\cline{2-7} 
 & ET & $38.34$ & $68.2$ & $150$ & $149.8$ & $21.36$ \tabularnewline
\cline{2-7} 
 & HN & $4.21\pm{0.95}$ & $2.66\pm{1.79}$ & $2.76\pm{0.47}$ & $10$ & $10$ \tabularnewline
\cline{2-7} 
 & NoP & $31.4\pm{6.77}$ & $52.75\pm{44.98}$ & $54.68\pm{10.92}$ & $72$ & $72$ \tabularnewline
 \hline
\end{tabular}
\par\end{centering}
\centering{}CR: classification rate, ET: execution time, HN: hidden nodes, NoP: number of parameters
\end{table*}
\section{Discriminative Training Phase}
Once the true class labels $C_k=[C_1,C_2,...,C_T]\in\Re^{T}$ are obtained, the 0-1 encoding scheme is applied to craft the target vector $C_k\in\Re^{T\times m}$ where $m$ is the number of target class. That is, $C_o=1$ if only if a data sample $X_t$ falls into $o$-th class. A generative model is passed to the discriminative training phase added with a softmax layer to infer the final classification decision as follows:
\begin{equation}
    \hat{C_t}=softmax(s(X_t W+b) \Phi+\eta)
\end{equation}
where $\Phi\in\Re^{R \times m}$ and $\eta\in\Re^{m}$ denote the output weight vector and bias of discriminative network respectively while the softmax layer outputs probability distribution across $m$ target classes $softmax(x_0)=\frac{exp(x_o)}{\sum_{k=1}^{m}exp(x_k)}$.

The parameters, $W,b,\Phi,\eta$ are further adjusted using the labelled data chunk $B_k=(X_k,C_k)\in\Re^{T\times(n+m)}$ via the SGD method with only a single epoch. The optimization problem is formulated as follows:
\begin{equation}
\arg\min_{W,b,\Phi,\eta}\sum_{t=1}^{T}\frac{1}{T}L(C_t,\hat{C}_t)   
\end{equation}
where the loss function is akin to the generative training phase, the cross-entropy loss function. The adjustment process is executed in the one-pass learning fashion leading to per-sample adaptation process $T=1$.

The structural learning scenario also occurs in the discriminative learning phase where the NS approach can be formulated in respect to the squared predictive error rather than reconstruction error $\frac{\sum_{t=1}^{T}(C_t-\hat{C}_t)^{2}}{T}$. Similar derivation can be applied here but the difference only exists in the output expression of the discrimininative model as $s(X_t W+b)\Phi+\eta$ instead of the encoding and decoding scheme as shown in (\ref{encoder}),(\ref{decoder}). Moreover, the hidden node growing and pruning conditions still refer to the same criteria (\ref{eq:HUgrowing}),(\ref{eq:HUpruning}).  The pseudocode of DEVDAN's generative and discriminative phases are placed in the supplemental document.   
\section{Proof of Concepts}
The learning performance of DEVDAN is numerically validated using nine real-world and synthetic data stream problems: SEA, Hyperplane, Susy, KDDCup, RLCPS, RFID localization, Hepmass, Electricity Pricing and Weather. At least five of nine problems characterize \textbf{non-stationary properties}, while the remainder four problems feature salient characteristics in examining the performance of data stream algorithms: big size, high input dimension, etc. We refer readers to supplemental document for detailed characteristics of the nine datasets including the number of time stamps applied in the prequential test-then-train procedure. The numerical results of DEVDAN is compared against conventional AE and DAE where the discriminative phase is adjusted using only a single training epoch to assure fair comparison. AE and DAE structures are initialized before process runs. Comparison against classic AE and DAE is shown to highlight to what extent DEVDAN outperforms its root while  DEVDAN is also compared against pENsemble \cite{pENsemble} and \cite{pensembleplus}- a prominent data stream algorithm built upon an evolving ensemble classifier concept. 

The learning performance of the consolidated algorithms is evaluated according to four criteria: classification rate, number of parameters, execution time and hidden units while the prequential test-then train procedure is followed as our evaluation protocol to simulate real data stream environments. The numerical results refer to the average numerical results across all time stamps. Numerical results are reported in Table 1. All consolidated algorithms are executed in the same computational platform under MATLAB environments with the Intel(R) Xeon(R) CPU E5-1650 @3.20 GHz processor and 16 GB RAM. Because of the page limit, all figures pertaining to DEVDAN learning performance and the source code of DEVDAN are placed as supplemental documents. The source code of DEVDAN will be made publicly available once our paper is accepted.
\subsection{Numerical Results}
It is reported in Table 1 that DEVDAN produces more accurate prediction than its counterparts in six problems: KDD Cup, SEA, Hyperplane, SUSY, RLCPS and HEPMASS. This fact confirms the efficacy of DEVDAN in coping with non-stationary learning environments because Hyperplane, SEA and KDD Cup problems are well-known in the literature for their non-stationary properties. DEVDAN consistently outperforms both AE and DAE having a fixed structure except only slightly inferior to DAE in the RFID localization problem. DEVDAN also exhibits very competitive performance against ensemble classifiers, pENsemble and pENsemble$+$. It is worth noting that pENsemble and pENsemble$+$ incurs much higher computational complexity than DEVDAN because it is crafted under the concept of multi-model structure. This fact is substantiated by the execution time of pENsemble and pENsemble$+$ consistently slower than DEVDAN in almost all problems. The learning performance of DEVDAN is visualized in the supplemental document.    
\section{Conclusion}
This paper presents a novel denoising autoencoder (DAE), namely the deep evolving denoising autoencoder (DEVDAN). DEVDAN features a self-organizing property in both generative and discriminative phases where input features can be incrementally constructed and discarded in a fully automated manner with the absence of a user-defined threshold. Our numerical study in nine popular data stream problems shows that DEVDAN delivers the most encouraging numerical result from other four benchmarked algorithms. Our numerical results demonstrate the advantage of DEVDAN's evolving structure which adapts to dynamic components of data streams. This fact also supports the relevance of generative phase for online data stream which contributes toward refinement of network structure in unsupervised fashion. Nevertheless, it is admitted that DEVDAN is still crafted under a single hidden layer feedforward network. A deep version of DEVDAN will be subject to our future investigation.    
\bibliographystyle{aaai}
\bibliography{ICDM}

\begin{thebibliography}{}

\bibitem[\protect\citeauthoryear{Alvares and
  Salzmann}{2016}]{LearningTheNumber}
Alvares, J.~M., and Salzmann, M.
\newblock 2016.
\newblock Learning the number of neurons in deep networks.
\newblock In Lee, D.~D.; Sugiyama, M.; Luxburg, U.~V.; Guyon, I.; and Garnett,
  R., eds., {\em Advances in Neural Information Processing Systems 29}. Curran
  Associates, Inc.
\newblock  2270--2278.

\bibitem[\protect\citeauthoryear{Baldi, Sadowski, and
  Whiteson}{2014}]{Baldi2014SearchingFE}
Baldi, P.; Sadowski, P.~D.; and Whiteson, D.
\newblock 2014.
\newblock Searching for exotic particles in high-energy physics with deep
  learning.
\newblock {\em Nature communications} 5:4308.

\bibitem[\protect\citeauthoryear{Bengio \bgroup et al\mbox.\egroup
  }{2006}]{Bengio_Greedy}
Bengio, Y.; Lamblin, P.; Popovici, D.; and Larochelle, H.
\newblock 2006.
\newblock Greedy layer-wise training of deep networks.
\newblock In {\em Proceedings of the 19th International Conference on Neural
  Information Processing Systems}, NIPS'06,  153--160.
\newblock Cambridge, MA, USA: MIT Press.

\bibitem[\protect\citeauthoryear{Bengio, Courville, and
  Vincent}{2013}]{Bengio_2013}
Bengio, Y.; Courville, A.; and Vincent, P.
\newblock 2013.
\newblock Representation learning: A review and new perspectives.
\newblock {\em IEEE Trans. Pattern Anal. Mach. Intell.} 35(8):1798--1828.

\bibitem[\protect\citeauthoryear{Bifet \bgroup et al\mbox.\egroup }{2010}]{MOA}
Bifet, A.; Holmes, G.; Kirkby, R.; and Pfahringer, B.
\newblock 2010.
\newblock Moa: Massive online analysis.
\newblock {\em J. Mach. Learn. Res.} 11:1601--1604.

\bibitem[\protect\citeauthoryear{Denil \bgroup et al\mbox.\egroup
  }{2013}]{parameterprediction}
Denil, M.; Shakibi, B.; Dinh, L.; Ranzato, M.; and de~Freitas, N.
\newblock 2013.
\newblock Predicting parameters in deep learning.
\newblock In {\em Proceedings of the 26th International Conference on Neural
  Information Processing Systems - Volume 2}, NIPS'13,  2148--2156.
\newblock USA: Curran Associates Inc.

\bibitem[\protect\citeauthoryear{Ditzler and Polikar}{2013}]{DitzlerImbalanced}
Ditzler, G., and Polikar, R.
\newblock 2013.
\newblock Incremental learning of concept drift from streaming imbalanced data.
\newblock {\em IEEE Trans. on Knowl. and Data Eng.} 25(10):2283--2301.

\bibitem[\protect\citeauthoryear{Gama \bgroup et al\mbox.\egroup
  }{2014}]{Gamaconceptdrift}
Gama, J.~a.; \v{Z}liobait\.{e}, I.; Bifet, A.; Pechenizkiy, M.; and Bouchachia,
  A.
\newblock 2014.
\newblock A survey on concept drift adaptation.
\newblock {\em ACM Comput. Surv.} 46(4):44:1--44:37.

\bibitem[\protect\citeauthoryear{Gama, Fernandes, and Rocha}{2006}]{Gama2006}
Gama, J.~a.; Fernandes, R.; and Rocha, R.
\newblock 2006.
\newblock Decision trees for mining data streams.
\newblock {\em Intell. Data Anal.} 10(1):23--45.

\bibitem[\protect\citeauthoryear{Gama}{2010}]{GamaDataStream}
Gama, J.
\newblock 2010.
\newblock {\em Knowledge Discovery from Data Streams}.
\newblock Chapman \& Hall/CRC, 1st edition.

\bibitem[\protect\citeauthoryear{Hinton and Salakhutdinov}{2006}]{HinSal06}
Hinton, G., and Salakhutdinov, R.
\newblock 2006.
\newblock Reducing the dimensionality of data with neural networks.
\newblock {\em Science} 313(5786):504 -- 507.

\bibitem[\protect\citeauthoryear{Hinton and Zemel}{1993}]{Hinton_AE}
Hinton, G.~E., and Zemel, R.~S.
\newblock 1993.
\newblock Autoencoders, minimum description length and helmholtz free energy.
\newblock In {\em Proceedings of the 6th International Conference on Neural
  Information Processing Systems}, NIPS'93,  3--10.
\newblock San Francisco, CA, USA: Morgan Kaufmann Publishers Inc.

\bibitem[\protect\citeauthoryear{Hinton, Vinyals, and Dean}{}]{distilling}
Hinton, G.; Vinyals, O.; and Dean, J.
\newblock {Distilling the Knowledge in a Neural Network}.
\newblock {\em arXiv preprint arXiv:1503.02531}.

\bibitem[\protect\citeauthoryear{Mohammadi \bgroup et al\mbox.\egroup
  }{2017}]{DEEPIOT}
Mohammadi, M.; Al-Fuqaha, A.~I.; Sorour, S.; and Guizani, M.
\newblock 2017.
\newblock Deep learning for iot big data and streaming analytics: A survey.
\newblock {\em CoRR} abs/1712.04301.

\bibitem[\protect\citeauthoryear{Murphy}{2012}]{Murphy_Machine_Learning}
Murphy, K.~P.
\newblock 2012.
\newblock {\em Machine Learning: A Probabilistic Perspective}.
\newblock The MIT Press.

\bibitem[\protect\citeauthoryear{Platt}{1991}]{Platt}
Platt, J.
\newblock 1991.
\newblock A resource-allocating network for function interpolation.
\newblock {\em Neural Comput.} 3(2):213--225.

\bibitem[\protect\citeauthoryear{Pratama \bgroup et al\mbox.\egroup
  }{2017}]{pensembleplus}
Pratama, M.; Dimla, E.; Lughofer, E.; Pedrycz, W.; and Tjahjowidodo, T.
\newblock 2017.
\newblock Online tool condition monitoring based on parsimonious ensemble+.
\newblock {\em CoRR} abs/1711.01843.

\bibitem[\protect\citeauthoryear{Pratama, Pedrycz, and
  Lughofer}{2018}]{pENsemble}
Pratama, M.; Pedrycz, W.; and Lughofer, E.
\newblock 2018.
\newblock Evolving ensemble fuzzy classifier.
\newblock {\em IEEE Transactions on Fuzzy Systems}  1--1.

\bibitem[\protect\citeauthoryear{Sahoo \bgroup et al\mbox.\egroup
  }{2017}]{OnlineDeepLearning}
Sahoo, D.; Pham, Q.~D.; Lu, J.; and Hoi, S.~C.
\newblock 2017.
\newblock Online deep learning: Learning deep neural networks on the fly.
\newblock {\em arXiv preprint arXiv:1711.03705} abs/1711.03705.

\bibitem[\protect\citeauthoryear{Sariyar, Borg, and Pommerening}{2011}]{RLCPS}
Sariyar, M.; Borg, A.; and Pommerening, K.
\newblock 2011.
\newblock Controlling false match rates in record linkage using extreme value
  theory.
\newblock {\em Journal of Biomedical Informatics} 44(4):648--654.

\bibitem[\protect\citeauthoryear{Stolfo \bgroup et al\mbox.\egroup
  }{2000}]{KDDCup}
Stolfo, S.~J.; Fan, W.; Lee, W.; Prodromidis, A.; and Chan, P.~K.
\newblock 2000.
\newblock Cost-based modeling for fraud and intrusion detection: Results from
  the jam project.
\newblock In {\em In Proceedings of the 2000 DARPA Information Survivability
  Conference and Exposition},  130--144.
\newblock IEEE Computer Press.

\bibitem[\protect\citeauthoryear{Street and Kim}{2001}]{SEA}
Street, W.~N., and Kim, Y.-S.
\newblock 2001.
\newblock A streaming ensemble algorithm (sea) for large-scale classification.
\newblock In {\em Proceedings of the Seventh ACM SIGKDD International
  Conference on Knowledge Discovery and Data Mining}, KDD '01,  377--382.
\newblock New York, NY, USA: ACM.

\bibitem[\protect\citeauthoryear{Vincent \bgroup et al\mbox.\egroup
  }{2008}]{VincentDAE}
Vincent, P.; Larochelle, H.; Bengio, Y.; and Manzagol, P.-A.
\newblock 2008.
\newblock Extracting and composing robust features with denoising autoencoders.
\newblock In {\em Proceedings of the 25th International Conference on Machine
  Learning}, ICML '08,  1096--1103.
\newblock New York, NY, USA: ACM.

\bibitem[\protect\citeauthoryear{Wang and Li}{2017}]{SCN}
Wang, D., and Li, M.
\newblock 2017.
\newblock Stochastic configuration networks: Fundamentals and algorithms.
\newblock {\em IEEE transactions on cybernetics} 47(10):3466--3479.

\bibitem[\protect\citeauthoryear{Yingwei, Sundararajan, and
  Saratchandran}{1997}]{MRAN}
Yingwei, L.; Sundararajan, N.; and Saratchandran, P.
\newblock 1997.
\newblock A sequential learning scheme for function approximation using minimal
  radial basis function neural networks.
\newblock {\em Neural Comput.} 9(2):461--478.

\bibitem[\protect\citeauthoryear{Yoon \bgroup et al\mbox.\egroup
  }{2018}]{DeepExpandable}
Yoon, J.; Yang, E.; Lee, J.; and Hwang, S.~J.
\newblock 2018.
\newblock Lifelong learning with dynamically expandable networks.
\newblock ICLR.

\bibitem[\protect\citeauthoryear{Zhou, Sohn, and
  Lee}{2012}]{Zhou_incrementallearning}
Zhou, G.; Sohn, K.; and Lee, H.
\newblock 2012.
\newblock Online incremental feature learning with denoising autoencoders.
\newblock {\em Journal of Machine Learning Research} 22:1453--1461.

\end{thebibliography}
\end{document}